\documentclass[9pt]{article}
\usepackage{spconf,amsmath,graphicx,hyperref}
\usepackage{array}
\usepackage{xcolor}
\usepackage{algorithm}
\usepackage{algorithmic}
\usepackage{arydshln}
\usepackage{amssymb}
\usepackage{booktabs}
\usepackage{bm}
\usepackage{bbding}
\usepackage{pifont}
\usepackage{wasysym}
\usepackage{amssymb}
\usepackage{booktabs}
\usepackage{array}
\usepackage{tabularx} 
\usepackage{arydshln}
\usepackage{multirow}
\usepackage{algorithmic}
\usepackage{graphicx}
\usepackage{textcomp}
\usepackage{xcolor}
\usepackage{arydshln}
\usepackage{amssymb}
\usepackage{geometry}
\usepackage{cite}
\usepackage{enumitem}

\title{Physics and Data Driven Transformer-Mamba Framework for Flow Field}
\name{\centering Zhuo Zhang \textsuperscript{a, *}, Shun Zou \textsuperscript{b, *} \thanks{\textsuperscript{*} The authors contribute equally, $\dagger$ Corresponding authors}, Canqun Yang \textsuperscript{a, d}, Xi Yang \textsuperscript{a, c, $\dagger$}\thanks{This work was partially supported by the National Key R\&D Program of China (Grant No.2023YFB3309100, 2023YFB3309105).}}
\address{\small
\centering
\begin{tabular}{c}
\textsuperscript{a} College of Computer Science and Technology, 
National University of Defense Technology, Changsha, China\\
\textsuperscript{b} College of Artificial Intelligence, Nanjing Agricultural University, NanJing, China\\
\textsuperscript{c} National Key Laboratory of Parallel and Distributed Computing, National University of Defense Technology, Changsha, China\\
\textsuperscript{d} National SuperComputer Center in Tianjin, Binhai New Area, Tianjin, China
\end{tabular}
}
\begin{document}
%
\maketitle
\begin{abstract}
While deep learning accelerates expensive partial differential equation solving in computational fluid dynamics (CFD), existing methods like PINNs and FNOs often struggle with generalization, noise robustness, and physical consistency. We introduce the Transformer-Mamba for Flow Field (TM4FF) framework, a physics-constrained operator learning model with three key innovations: a Residual Wavelet Mamba (RWM) layer for feature denoising, a Transformer-based attention mechanism for enhanced feature fusion, and a physics-informed loss using Fourier derivatives to enforce the Navier-Stokes equations. Experiments on four CFD datasets show TM4FF achieves high accuracy and robust generalization across varying flow conditions.
\end{abstract}

\section{Introduction}
\vspace{-1mm} 

\begin{figure*}[t]
\centering
\includegraphics[width=0.85\linewidth]{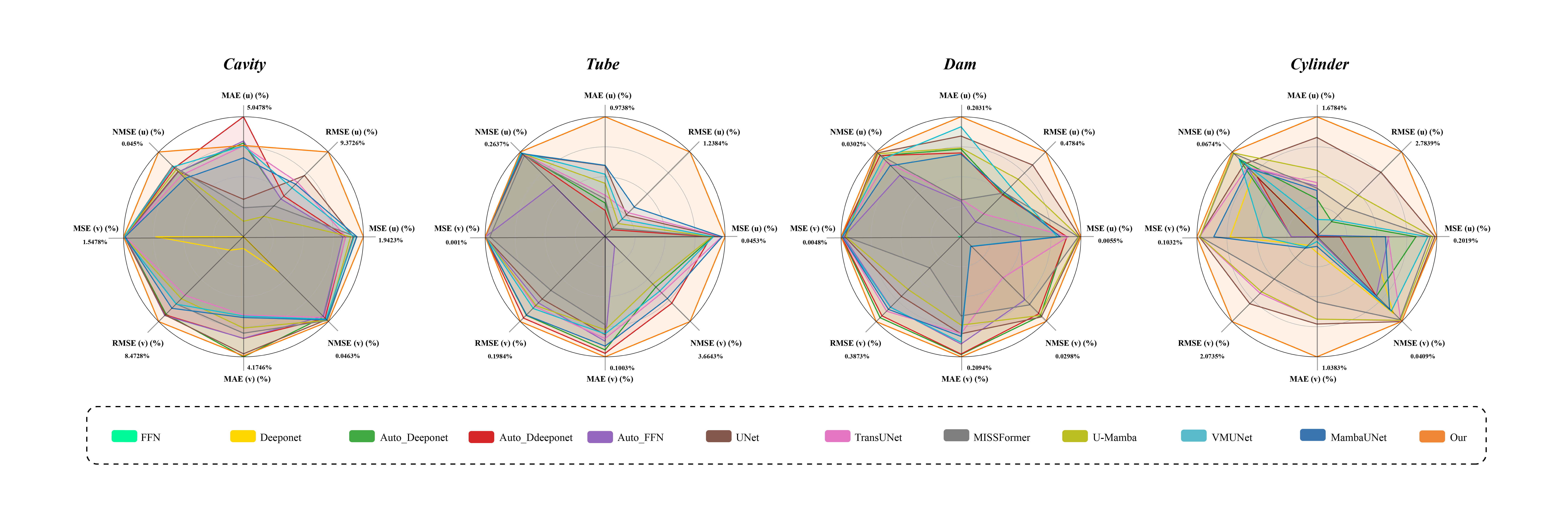}
\vspace{-0.5cm}
\caption{Overall evaluation metrics for each method in flow field prediction.}
\label{leida}
\end{figure*}

\begin{figure*}[t]
	\centering
	\includegraphics[width=0.8\linewidth]{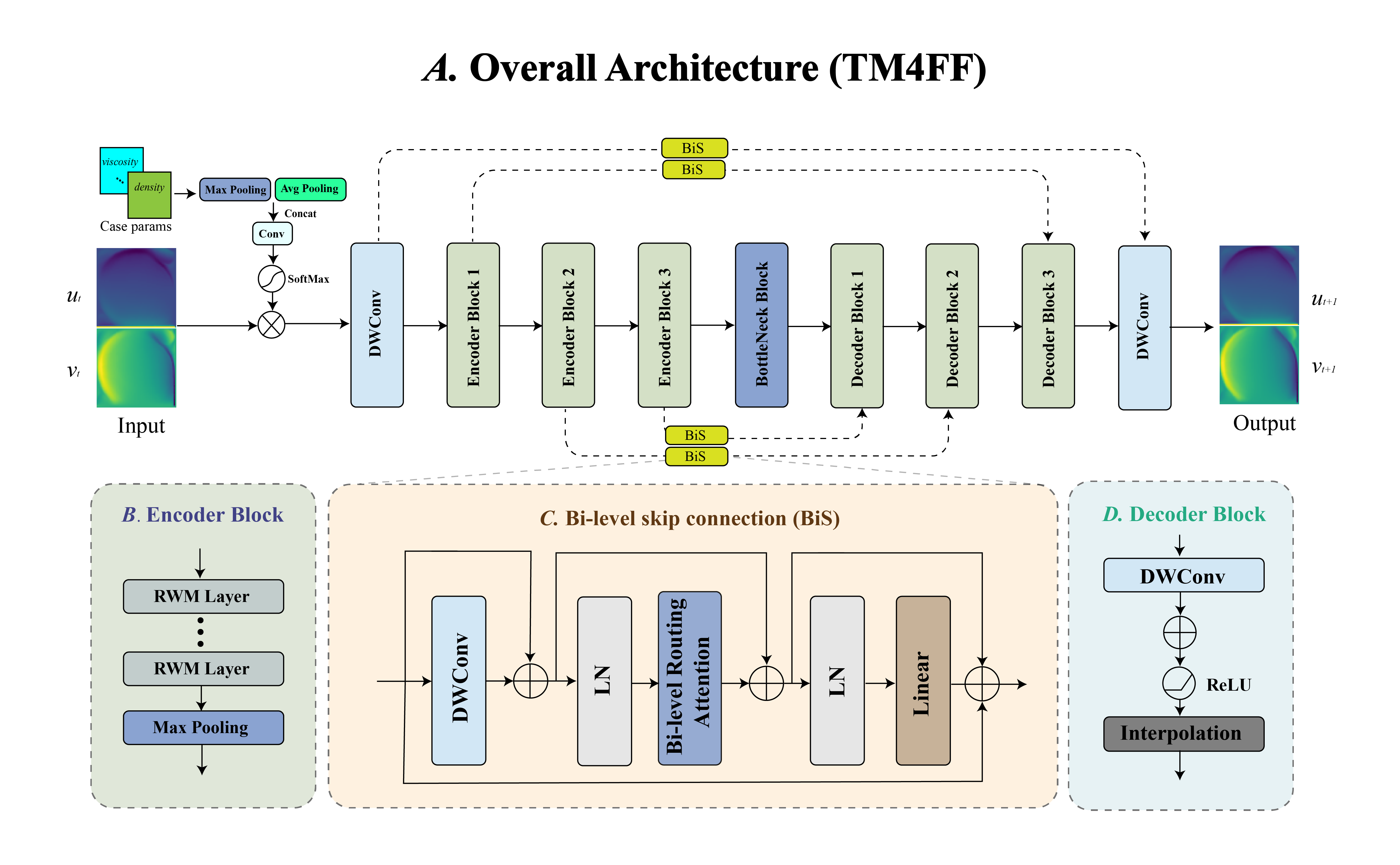}
    \vspace{-0.7cm}
	\caption{Overview of Transformer-Mamba for Flow Field (TM4FF). RWM: Residual Wavelet Mamba, DWConv: Depthwise Convolution, LN: LayerNorm.}
    \vspace{-0.2cm}
	\label{net}
\end{figure*}

\begin{figure*}[t]
	\centering
	\includegraphics[width=0.8\linewidth]{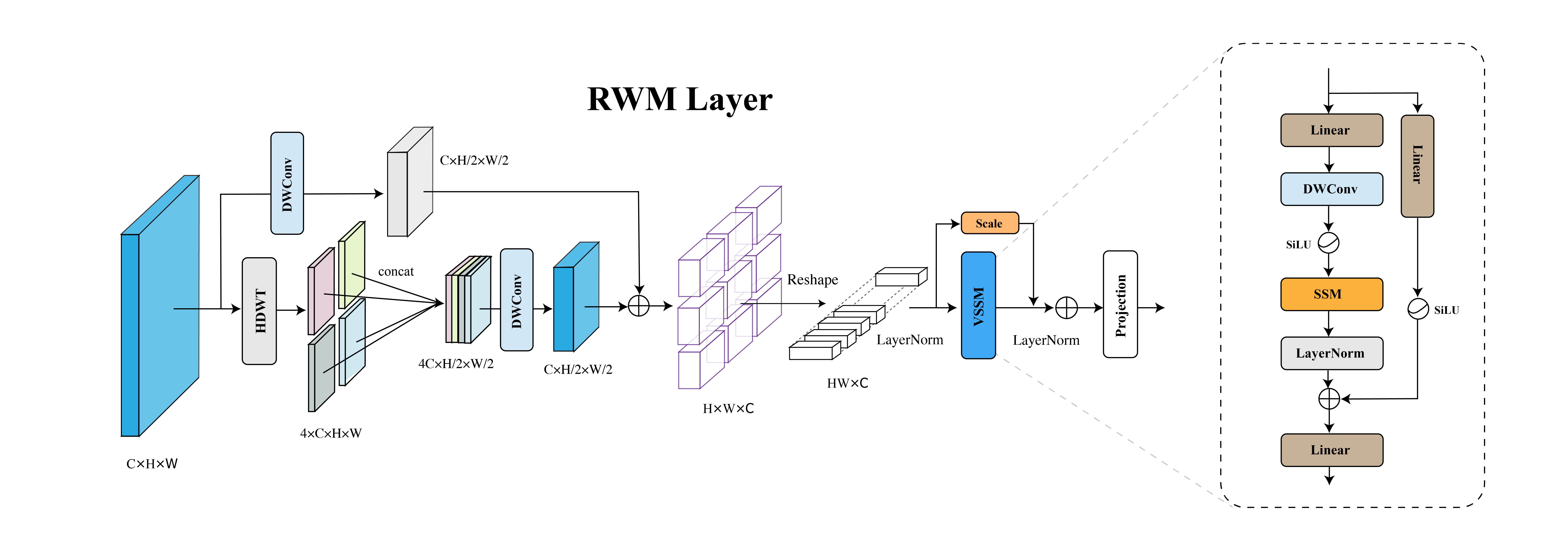}
    \vspace{-0.7cm}
	\caption{Overview of the Residual Wavelet Mamba (RWM) layer, containing the core Vision State-Space Module (VSSM).}
	\label{rwm}
    \vspace{-0.3cm}
\end{figure*}
Solving partial differential equations (PDEs) in computational fluid dynamics (CFD) with traditional numerical methods \cite{fdmsmith1985numerical, fvmeymard2000finite, femzienkiewicz2005finite} is computationally intensive. Deep learning offers a promising alternative to accelerate simulations by orders of magnitude \cite{icassp1DBLP:conf/icassp/ParkGYS25, nips3DBLP:conf/nips/DangelMZ24}, with some models already surpassing traditional methods in accuracy for certain problems \cite{aaai5yang2023dmis, icassp2DBLP:conf/icassp/Zhang0025, nudt1yan2023st, nudtxinhaichen2024neural, me2zhang2025legend}.

Two dominant deep learning paradigms have emerged to tackle these problems. Physics-Informed Neural Networks (PINNs) \cite{pinnsraissi2019physics, mupinnzhang2023multi, me} integrate physical laws as constraints but often struggle to generalize across varying flow conditions \cite{icml1DBLP:conf/icml/SalehGBO024, icml2DBLP:conf/icml/ChoJLL00P24, nips1DBLP:conf/nips/HwangL24}. In response, data-driven operator learning methods, such as Fourier Neural Operators (FNOs) \cite{fnoli2020fourier} and Deep Operator Networks (DeepONets) \cite{deooponetlu2021learning}, were developed to learn mappings between function spaces, offering better generalization. While these models, including architectures like U-Net \cite{unetronneberger2015u}, successfully capture spatial hierarchies, several challenges remain:

\textbf{\textit{W1:)}} Difficulty in modeling the complex temporal dependencies and multi-scale correlations present in unsteady flow fields.

\textbf{\textit{W2:)}} Poor generalization to unseen PDE parameters, physical properties, domain geometry, etc.), requiring retraining as parameters change.

\textbf{\textit{W3:)}} Current data-driven methods lack explicit physical constraints. Although networks learn certain physical laws through training, these are often implicit and may not guarantee adherence to fundamental physical principles.

\textbf{\textit{W4:)}} Noise generated during training (e.g., striping noise) affects the accuracy of specific flow field predictions.

To address these limitations, we propose the Transformer-Mamba for Flow Field (TM4FF) framework. TM4FF treats fluid flow as a spatiotemporal image sequence and enhances generalization by directly embedding operational parameters into the neural network. We introduce several architectural innovations: a novel Residual Wavelet Mamba (RWM) Layer uses a dual-branch structure with Haar DWT to decouple and filter noise (addressing \textbf{\textit{W4}}). We replace traditional skip connections with a Transformer-based self-attention mechanism to intelligently fuse multi-scale features (addressing \textbf{\textit{W2}}). Finally, we incorporate a physics-informed loss function with Fourier derivatives to enforce the Navier-Stokes equations, ensuring the model's predictions are physically consistent (addressing \textbf{\textit{W3}}), while the Mamba architecture itself excels at capturing long-range spatiotemporal dependencies (addressing \textbf{\textit{W1}}). As shown in Fig.~\ref{leida}, comprehensive experiments demonstrate that our proposed method achieves state-of-the-art performance.

Our main contributions are:
\begin{itemize}[nosep, leftmargin=*]
    \item \textbf{TM4FF Framework:} A novel architecture that embeds operational parameters and uses a Mamba encoder to capture spatiotemporal characteristics of unsteady flow.
    \item \textbf{Residual Wavelet Mamba Layer:} A dual-branch layer for noise decoupling, combining a model-driven Haar DWT branch with a data-driven residual branch.
    \item \textbf{Transformer-based Skip Connection:} An intelligent feature fusion mechanism that replaces standard skip connections to enhance information flow between the encoder and decoder.
    \item \textbf{Physics-Informed Loss with Fourier Derivatives:} A hybrid loss function that constrains the data-driven model with the Navier-Stokes equations for improved accuracy and physical realism.
\end{itemize}

\begin{table*}[h!]
\scriptsize 
\caption{Comparative Study of Fluid Prediction Methods and SOTA Models (Including Autoregressive and Non-Autoregressive Models). The bold black values represent the best results, while the underlined values indicate the second-best results.}
\label{tab:efficiency_comparison}
\centering
\tiny
\begin{tabular}{lccccccccccc}
\toprule
\textbf{Case} & \textbf{Model} & \textbf{Source} & \textbf{Param (K)} & \textbf{MSE (u)} & \textbf{RMSE (u)} & \textbf{MAE (u)} & \textbf{NMSE (u)} & \textbf{MSE (v)} & \textbf{RMSE (v)} & \textbf{MAE (v)} & \textbf{NMSE (v)} \\
\midrule
\multirow{12}{*}{Cavity} & FFN & JCP-2019 & 72 & 7.5073394 & 2.4209118 & 1.1220862 & 0.1134040 & 113.7456860 & 9.3089929 & 7.1327300 & 1.4379886 \\
 & Deeponet & NMI-2019 & 143 & 1.9921126 & 1.2580220 & 0.7409260 & 0.0416701 & 1.1387197 & 0.6104441 & 0.5477043 & 0.0400961 \\
 & Auto\_Deeponet & NMI-2019 & 552 & 0.0639269 & 0.1530177 & 0.0587990 & 0.0013988 & 0.0662839 & 0.1114433 & \textbf{0.0417463} & 0.0030466 \\
 & Auto\_Edeeponet & Arxiv-2022 & 623 & 0.0571324 & 0.1476592 & \textbf{0.0504779} & 0.0011877 & 0.0456277 & 0.1088518 & 0.0788765 & 0.0022441 \\
 & Auto\_FFN & JCP-2019 & 1,102 & 0.0639877 & 0.1531136 & \underline{0.0581093} & 0.0014016 & 0.0733171 & \underline{0.1063512} & 0.0792814 & 0.0033839 \\
 & UNet & MICCAI-2015 & 1,095 & 0.0404604 & \underline{0.1178060} & 0.0894393 & 0.0013797 & 0.0436799 & 0.1080112 & 0.0475035 & 0.0014662 \\
 & TransUNet & MIA-2024 & 10,312 & 0.0568182 & 0.1264915 & 0.0598381 & 0.0015820 & 0.0697967 & 0.1926272 & 0.1372103 & 0.0028031 \\
 & MISSFormer & TMI2023 & 42,432 & 0.0336361 & 0.1675146 & 0.0978472 & \underline{0.0010864} & 0.0298596 & 0.1525197 & 0.0908791 & \underline{0.0010184} \\
 & U-Mamba & Arxiv-2024 & 12,143 & 0.0462004 & 0.1970269 & 0.1181226 & 0.0012765 & 0.0343462 & 0.1760469 & 0.1030626 & 0.0011706 \\
 & VMUNet & Arxiv-2024 & 63,321 & 0.0384606 & 0.1343874 & 0.0598663 & 0.0011342 & \underline{0.0280631} & 0.1520837 & 0.1345357 & 0.0017958 \\
 & MambaUNet & Arxiv-2024 & 58,536 & \underline{0.0326497} & 0.1291396 & 0.0646101 & 0.0017808 & 0.0651264 & 0.1344283 & 0.1303377 & 0.0013429 \\
 & Ours & - & 2,479 & \textbf{0.0194232} & \textbf{0.0937261} & 0.0597321 & \textbf{0.0004501} & \textbf{0.0154782} & \textbf{0.0847283} & \underline{0.0436231} & \textbf{0.0004632} \\
\midrule
\multirow{12}{*}{Tube} & FFN & JCP-2019 & 72 & 0.6707547 & 0.4613486 & 0.3809852 & 5.2505060 & 2.2916732 & 1.1345139 & 0.9788251 & 1.0024409 \\
 & Deeponet & NMI-2019 & 143 & 0.6128914 & 0.4384824 & 0.3820198 & 7.1046984 & 0.6372135 & 0.4460260 & 0.3856814 & 4.2262289 \\
 & Auto\_Deeponet & NMI-2019 & 552 & 0.0027213 & 0.0440116 & 0.0244697 & 0.0051395 & 0.0000260 & 0.0046837 & 0.0028324 & 0.1050158 \\
 & Auto\_Edeeponet & Arxiv-2022 & 623 & 0.0030899 & 0.0467656 & 0.0273037 & 0.0067682 & \underline{0.0000141} & \underline{0.0034561} & \underline{0.0019282} & \underline{0.0687442} \\
 & Auto\_FFN & JCP-2019 & 1,102 & 0.1635600 & 0.2396413 & 0.1549187 & 0.1136883 & 0.0001718 & 0.0107982 & 0.0052860 & 0.3184348 \\
 & UNet & MICCAI-2015 & 1,095 & 0.0014302 & 0.0315242 & 0.0158496 & 0.0092400 & 0.0002091 & 0.0128815 & 0.0072763 & 0.8880437 \\
 & TransUNet & MIA-2024 & 10,312 & 0.0010554 & 0.0301301 & 0.0220797 & \underline{0.0045237} & 0.0003843 & 0.0089396 & 0.0060865 & 0.0906755 \\
 & MISSFormer & TMI2023 & 42,432 & 0.0021123 & 0.0442941 & 0.0231487 & 0.0131996 & 0.0029621 & 0.0181901 & 0.0106024 & 1.3152490 \\
 & U-Mamba & Arxiv-2024 & 12,143 & 0.0029736 & 0.0382078 & 0.0191961 & 0.0059186 & 0.0001042 & 0.0094126 & 0.0088644 & 0.1130801 \\
 & VMUNet & Arxiv-2024 & 63,321 & 0.0026259 & 0.0344924 & 0.0173475 & 0.0049388 & 0.0000895 & 0.0079654 & 0.0075437 & 0.0952159 \\
 & MambaUNet & Arxiv-2024 & 58,536 & \underline{0.0009713} & \underline{0.0272554} & \underline{0.0158096} & 0.0069618 & 0.0000853 & 0.0048083 & 0.0038873 & 0.0781457 \\
 & Ours & - & 2,479 & \textbf{0.0004528} & \textbf{0.0123843} & \textbf{0.0097382} & \textbf{0.0026371} & \textbf{0.0000101} & \textbf{0.0019837} & \textbf{0.0010031} & \textbf{0.0366431} \\
\midrule
\multirow{12}{*}{Dam} & FFN & JCP-2019 & 72 & 0.0764840 & 0.2295431 & 0.1841930 & 0.2973458 & 0.0630625 & 0.2071351 & 0.1587659 & 0.2642470 \\
 & Deeponet & NMI-2019 & 143 & 0.0752924 & 0.2224116 & 0.1786302 & 0.2955446 & 0.5406630 & 0.7121797 & 0.5571606 & 1.1806147 \\
 & Auto\_Deeponet & NMI-2019 & 552 & 0.0003602 & 0.0096556 & 0.0038370 & 0.0007205 & \underline{0.0000952} & \underline{0.0049292} & \underline{0.0024394} & 0.0028124 \\
 & Auto\_Edeeponet & Arxiv-2022 & 623 & 0.0003540 & 0.0097928 & 0.0040903 & 0.0007160 & 0.0001253 & 0.0054513 & 0.0025079 & 0.0037854 \\
 & Auto\_FFN & JCP-2019 & 1,102 & 0.0017431 & 0.0169537 & 0.0089755 & 0.0033383 & 0.0004344 & 0.0080703 & 0.0040738 & 0.0111886 \\
 & UNet & MICCAI-2015 & 1,095 & \underline{0.0000777} & \underline{0.0059747} & 0.0030416 & \underline{0.0004109} & 0.0006042 & 0.0117739 & 0.0058359 & \underline{0.0023922} \\
 & TransUNet & MIA-2024 & 10,312 & 0.0003259 & 0.0139952 & 0.0086372 & 0.0014553 & 0.0001962 & 0.0067997 & 0.0058663 & 0.0269536 \\
 & MISSFormer & TMI2023 & 42,432 & 0.0001088 & 0.0094345 & 0.0087492 & 0.0005011 & 0.0007822 & 0.0261032 & 0.0093841 & 0.0084725 \\
 & U-Mamba & Arxiv-2024 & 12,143 & 0.0000949 & 0.0075590 & 0.0037587 & 0.0004991 & 0.0007619 & 0.0144648 & 0.0074682 & 0.0030714 \\
 & VMUNet & Arxiv-2024 & 63,321 & 0.0005829 & 0.0085537 & \underline{0.0025350} & 0.0011020 & 0.0001998 & 0.0079952 & 0.0043473 & 0.0795341 \\
 & MambaUNet & Arxiv-2024 & 58,536 & 0.0005130 & 0.0094484 & 0.0041801 & 0.0019713 & 0.0002363 & 0.0073668 & 0.0053607 & 0.0787073 \\
 & Ours & - & 2,479 & \textbf{0.0000548} & \textbf{0.0047839} & \textbf{0.0020312} & \textbf{0.0003018} & \textbf{0.0000483} & \textbf{0.0038726} & \textbf{0.0020938} & \textbf{0.0002978} \\
\midrule
\multirow{12}{*}{Cylinder} & FFN & JCP-2019 & 72 & 0.4540003 & 0.3018296 & 0.2465903 & 0.3831972 & 0.6024798 & 0.3679617 & 0.2710318 & 0.4952313 \\
 & Deeponet & NMI-2019 & 143 & 0.0410685 & 0.1592126 & 0.1055810 & 0.0128104 & 0.0223351 & 0.0999994 & 0.0635683 & 0.0059943 \\
 & Auto\_Deeponet & NMI-2019 & 552 & 0.0115449 & 0.0762912 & 0.0445775 & 0.0043905 & 0.0969724 & 0.2010580 & 0.1168046 & 0.0195839 \\
 & Auto\_Edeeponet & Arxiv-2022 & 623 & 0.0783936 & 0.1867794 & 0.1143256 & 0.0155843 & 0.0966148 & 0.2022386 & 0.1177360 & 0.0194570 \\
 & Auto\_FFN & JCP-2019 & 1,102 & 0.0271349 & 0.1279762 & 0.0378661 & 0.0089120 & 0.0972226 & 0.2027037 & 0.1178205 & 0.0197800 \\
 & UNet & MICCAI-2015 & 1,095 & \underline{0.0027909} & \underline{0.0361314} & \underline{0.0215571} & 0.0071332 & \underline{0.0023470} & \underline{0.0297018} & \underline{0.0194579} & \textbf{0.0004094} \\
 & TransUNet & MIA-2024 & 10,312 & 0.0274852 & 0.1279372 & 0.0363029 & 0.0090431 & 0.0025086 & 0.0360006 & 0.0210826 & 0.0009956 \\
 & MISSFormer & TMI2023 & 42,432 & 0.0048139 & 0.0590244 & 0.0388507 & 0.0014658 & 0.0028199 & 0.0465320 & 0.0275426 & 0.0016540 \\
 & U-Mamba & Arxiv-2024 & 12,143 & 0.0037019 & 0.0489709 & 0.0313942 & \underline{0.0011376} & 0.0025853 & 0.0369812 & 0.0210547 & 0.0012535 \\
 & VMUNet & Arxiv-2024 & 63,321 & 0.0059625 & 0.0733041 & 0.0618380 & 0.0043218 & 0.0523000 & 0.0940000 & 0.0900400 & 0.0075901 \\
 & MambaUNet & Arxiv-2024 & 58,536 & 0.0293402 & 0.1329643 & 0.0400001 & 0.0095561 & 0.0110575 & 0.0911070 & 0.0749794 & 0.0088076 \\
 & Ours & - & 2,479 & \textbf{0.0020187} & \textbf{0.0278392} & \textbf{0.0167836} & \textbf{0.0006736} & \textbf{0.0010321} & \textbf{0.0207352} & \textbf{0.0103826} & \underline{0.0005091} \\
\bottomrule
\end{tabular}
\end{table*}

\section{Methodology}
\label{sec:method}
\vspace{-1mm}

\subsection{Overall Framework and Condition Embedding}
\vspace{-1mm} 

We aim to approximate the flow field in a domain $\mathcal{D} = \{(x, y, t)\}$. The TM4FF framework (Fig.~\ref{net}(A)) learns an operator $G: (\Sigma, \Omega) \mapsto u$, where $\Omega$ represents operational parameters (e.g., boundary conditions, viscosity), $\Sigma$ is the velocity field at time $t-\Delta t$, and $u$ is the field at time $t$. The network $f_\theta$ approximates this operator, with optimal parameters $\theta^*$ found by minimizing a loss function $\mathcal{L}(u, \hat{u})$.

To make the model aware of varying conditions, operational parameters $\Omega$ are embedded. Scalar parameters are tiled to match the spatial resolution of the input flow field $\mathbf{u}$ and concatenated into a tensor $\mathbf{x}$. Global pooling and a fully connected (FC) layer on $\mathbf{x}$ produce a feature vector $\mathbf{h}$:
\vspace{-2mm}
\begin{equation}
\mathbf{h} = \text{FC}(\text{concat}(\text{Globalmaxpool}(\mathbf{x}), \text{Globalavgpool}(\mathbf{x})))
\vspace{-1mm}
\end{equation}

A Sigmoid function generates attention weights $\mathbf{a} = \sigma(\mathbf{h})$, which modulate the input flow field: $\mathbf{u}_{\text{in}} = \mathbf{a} \odot \mathbf{u}$. This allows the model to adaptively respond to the given physical conditions.

\subsection{Residual Wavelet Mamba Encoder}
\vspace{-1mm} 

\begin{figure}[t]
	\centering
	\includegraphics[width=0.7\linewidth]{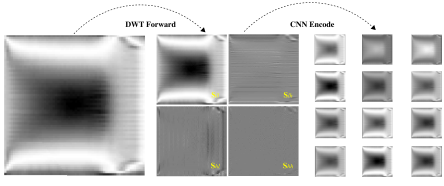}
    \vspace{-0.5cm}
	\caption{The wavelet decomposition and reconstruction of strip features in neural networks (DWT aggregates strip noise into \(S_{ll}\) and \(S_{hl}\)).
}

	\label{xiaobo}
\end{figure}

The TM4FF encoder (Fig.~\ref{net}(B)) uses blocks containing our novel Residual Wavelet Mamba (RWM) layer (Fig.~\ref{rwm}) for feature extraction and noise decoupling. The RWM layer processes an input feature map $S \in \mathbb{R}^{C \times H \times W}$ using a dual-branch structure. A model-driven branch first applies a 2D Haar Wavelet Transform (DWT) to decompose $S$ into four sub-bands (as shown in Fig.~\ref{xiaobo}):
\setlength{\abovedisplayskip}{1pt}%
\setlength{\belowdisplayskip}{1pt}%
\begin{equation}
S_{a} = \lambda_a \ast S, \quad a \in \{ll, lh, hl, hh\}
\end{equation}%

where $\lambda_a$ are Haar filters. This isolates features by frequency, helping to filter structured noise. These sub-bands are then processed and fused with features from a parallel data-driven residual branch. The fused output is passed to a Visual State Space Module (VSSM) \cite{vssmzhu2024vision}, which efficiently captures long-range spatial dependencies. The VSSM block processes input $W_{\text{in}}$ via two parallel branches:
\begin{align}
W_1 &= \operatorname{LN}(\operatorname{SSM}(\operatorname{SiLU}(\operatorname{DWConv}(\operatorname{Linear}(W_{\text{in}}))))) \\
W_2 &= \operatorname{SiLU}(\operatorname{Linear}(W_{\text{in}})) \\
W_{\text{out}} &= \operatorname{Linear}(W_1 \odot W_2)
\end{align}
This design combines physical priors (from wavelets) with the powerful sequence modeling capabilities of Mamba.

\subsection{Bi-level Skip Connection (BiS)}
\vspace{-1mm} 

\begin{figure}[hbtp]
	\centering
	\includegraphics[width=0.7\linewidth]{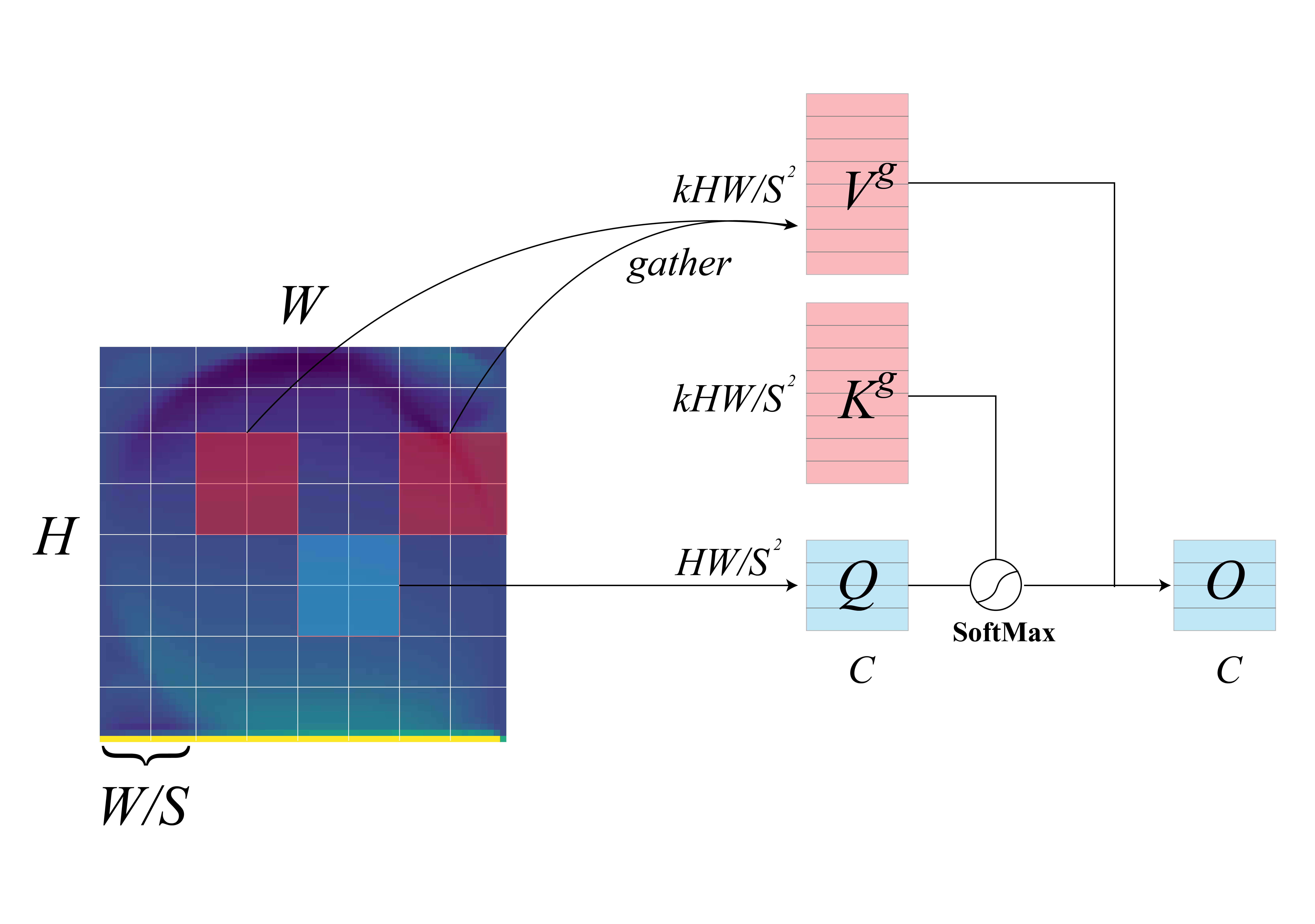}
  \vspace{-0.7cm}
	\caption{Overview of Bi-level Routing Attention.}
	\label{att}
  \vspace{-0.5cm}
\end{figure}
We replace standard skip connections with a Bi-level skip connection (BiS) to improve encoder-decoder feature fusion. This module employs Bi-level Routing Attention (as shown in Fig.~\ref{att}) \cite{biformerzhu2023biformer}, a sparse attention mechanism. For a feature map divided into regions, it computes an adjacency matrix $\mathbf{A}^r$ between region-level queries and keys. Instead of dense attention, it finds the top-k most relevant regions for each query:
\begin{equation}
\mathbf{I}^r = \operatorname{topkIndex}(\mathbf{A}^r)
\end{equation}
Attention is then computed only over these selected regions. The final output combines this sparse attention with a Local Context Enhancement (LCE) operation \cite{lvmren2022shunted}:
\begin{equation}
\mathbf{O} = \operatorname{Attention}(\mathbf{Q}, \mathbf{K}^g, \mathbf{V}^g) + \operatorname{LCE}(\mathbf{V})
\end{equation}

\subsection{Physics-Informed Loss Function}
\vspace{-1mm}
To ensure physical plausibility, we constrain the model with the incompressible Navier-Stokes (N-S) equations via a hybrid loss function. The total loss $\mathcal{L} = \mathcal{L}_{\text{data}} + \mathcal{L}_{\text{incompressible}} + \mathcal{L}_{\text{NS}}$ is a weighted sum of the standard MSE data loss ($\mathcal{L}_{\text{data}}$) and two physics-based penalties. These penalties are computed efficiently using Fourier Derivatives and consist of the incompressibility loss, which enforces the divergence-free condition $\mathcal{L}_{\text{incompressible}} = \left\| \nabla \cdot \mathbf{u} \right\|^2$, and the Navier-Stokes residual loss, which penalizes deviations from the momentum equation:
\begin{equation}
    \mathcal{L}_{\text{NS}} = \left\| \frac{\partial \mathbf{u}}{\partial t} + (\mathbf{u} \cdot \nabla) \mathbf{u} - \nu \nabla^2 \mathbf{u} - \mathbf{f} \right\|^2
\end{equation}
This formulation guides the network towards solutions that are both accurate and physically consistent.
\section{Experimental Results}
\vspace{-1mm} 
\subsection{Experimental Setup}

We benchmarked our method on four diverse fluid dynamics datasets from the CFDBench \cite{luo2023cfdbench}: lid-driven cavity, circular tube, breaking dam, and flow around a cylinder. The data was split into 8:1:1 training, validation, and test sets, ensuring disjoint operating parameters between splits. All models were trained for 100 epochs on an NVIDIA GeForce RTX 3090, with a learning rate decay factor of 0.9 applied every 20 epochs.

\begin{table}[ht] 
\centering
\caption{Ablation Study of Various Strategies}
\scriptsize
\setlength{\tabcolsep}{5pt}
\begin{tabular}{lcccc}
\toprule
\textbf{Components} & \textbf{Cavity} & \textbf{Tube} & \textbf{Dam} & \textbf{Cylinder} \\
\midrule
Backbone only & 0.0404604 & 0.0014302 & 0.0000777 & 0.0027909 \\
+ Case Params & 0.0389877 & 0.0012312 & 0.0014455 & 0.0027534 \\
+ Mamba       & 0.0397174 & 0.0013258 & 0.0007616 & 0.0027721 \\
+ BiS         & 0.0392813 & 0.0012937 & 0.0011103 & 0.0027408 \\
+ Params + BiS& 0.0382510 & 0.0010783 & 0.0010610 & 0.0026881 \\
+ Mamba + BiS & 0.0390156 & 0.0011826 & 0.0007944 & 0.0027615 \\
\midrule
\textbf{Ours (Full Model)} & \textbf{0.0194232} & \textbf{0.0004528} & \textbf{0.0000548} & \textbf{0.0020187} \\
\bottomrule
\end{tabular}
\label{tab:ablation_module}
\end{table}

\begin{table}[htbp]
\vspace{-0.3cm}
\centering
\caption{Ablation Study of Residual Wavelet Mamba on Model Performance}
\small
\begin{tabular}{ccccc}
\toprule
RWM & Cavity & Tube & Dam & Cylinder \\
\midrule
\XSolidBrush  & 0.0226239 & 0.0005742 & 0.0000667 & 0.0025262 \\
\Checkmark  &  \textbf{0.0194232} & \textbf{0.0004528} & \textbf{0.0000548} & \textbf{0.0020187} \\
\toprule 
\end{tabular}
\label{tab:ablation_rwm}
\vspace{-0.3cm}
\end{table}

\subsection{Results}
\vspace{-1mm} 

A comprehensive quantitative comparison, provided in Table~\ref{tab:efficiency_comparison}, demonstrates our method's superior performance against a suite of autoregressive and non-autoregressive models, including FFN \cite{pinnsraissi2019physics}, UNet \cite{unetronneberger2015u}, TransUNet \cite{transunetchen2021transunet}, and various Mamba-based architectures \cite{ma2024u, vmunetruan2024vm, mambaunetwang2024mamba}. Across all four fluid flow cases (Cavity, Tube, Dam, and Cylinder), our model consistently achieves state-of-the-art results, securing the lowest error in nearly every metric for both $u$ and $v$ velocity components. For instance, in the complex Cylinder case, our method significantly surpasses strong baselines like UNet. This high accuracy is achieved while maintaining superior parameter efficiency compared to larger models like MISSFormer and VMUNet. The overall results confirm that our approach is not only highly accurate but also efficient and robust across diverse fluid dynamics scenarios.

\subsection{Ablation Study}
\vspace{-1mm} 
We conducted ablation studies to systematically validate the contribution of each component of our model. The results, shown in Table~\ref{tab:ablation_module} and Table~\ref{tab:ablation_rwm}, confirm their effectiveness. Table~\ref{tab:ablation_module} demonstrates that integrating case parameter embedding, the Mamba encoder, and the BiS skip connections progressively improves performance over the baseline backbone, with the full model achieving the best results. Furthermore, Table~\ref{tab:ablation_rwm} isolates the impact of our novel Residual Wavelet Mamba (RWM) layer, showing that its inclusion provides a substantial performance boost across all datasets. For example, enabling RWM reduces the MSE in the Cavity case from 0.0226 to 0.0194. These studies verify that each proposed component is crucial to the model's overall success.

\section{Conclusion}
\label{sec:conclusion}
\vspace{-1mm} 

We propose the Transformer-Mamba for Flow Field framework, a model that enhances computational fluid dynamics prediction by embedding operational parameters. Its key innovations include a Residual Wavelet Mamba layer for feature denoising, a Transformer-based attention mechanism to replace skip connections, and a physics-informed loss with Fourier derivatives to enforce Navier-Stokes equation constraints. Experimental results show that TM4FF achieves high accuracy and generalization across varying flow conditions, providing a promising alternative to traditional CFD methods.

\bibliographystyle{IEEEbib}
\small\bibliography{new}

@article{fnoli2020fourier, title={Fourier neural operator for parametric partial differential equations}, author={Li, Zongyi and Kovachki, Nikola and Azizzadenesheli, Kamyar and Liu, Burigede and Bhattacharya, Kaushik and Stuart, Andrew and Anandkumar, Anima}, journal={arXiv preprint arXiv:2010.08895}, year={2020} }

@article{deooponetlu2021learning, title={Learning nonlinear operators via DeepONet based on the universal approximation theorem of operators}, author={Lu, Lu and Jin, Pengzhan and Pang, Guofei and Zhang, Zhongqiang and Karniadakis, George Em}, journal={Nature machine intelligence}, volume={3}, number={3}, pages={218--229}, year={2021}, publisher={Nature Publishing Group UK London} }

@inproceedings{unetronneberger2015u, title={U-net: Convolutional networks for biomedical image segmentation}, author={Ronneberger, Olaf and Fischer, Philipp and Brox, Thomas}, booktitle={Medical image computing and computer-assisted intervention--MICCAI 2015: 18th international conference, Munich, Germany, October 5-9, 2015, proceedings, part III 18}, pages={234--241}, year={2015}, organization={Springer} }

@inproceedings{biformerzhu2023biformer, title={Biformer: Vision transformer with bi-level routing attention}, author={Zhu, Lei and Wang, Xinjiang and Ke, Zhanghan and Zhang, Wayne and Lau, Rynson WH}, booktitle={Proceedings of the IEEE/CVF conference on computer vision and pattern recognition}, pages={10323--10333}, year={2023} }

@article{vssmzhu2024vision, title={Vision mamba: Efficient visual representation learning with bidirectional state space model}, author={Zhu, Lianghui and Liao, Bencheng and Zhang, Qian and Wang, Xinlong and Liu, Wenyu and Wang, Xinggang}, journal={arXiv preprint arXiv:2401.09417}, year={2024} }

@inproceedings{lvmren2022shunted, title={Shunted self-attention via multi-scale token aggregation}, author={Ren, Sucheng and Zhou, Daquan and He, Shengfeng and Feng, Jiashi and Wang, Xinchao}, booktitle={Proceedings of the IEEE/CVF conference on computer vision and pattern recognition}, pages={10853--10862}, year={2022} }

@article{transunetchen2021transunet, title={Transunet: Transformers make strong encoders for medical image segmentation}, author={Chen, Jieneng and Lu, Yongyi and Yu, Qihang and Luo, Xiangde and Adeli, Ehsan and Wang, Yan and Lu, Le and Yuille, Alan L and Zhou, Yuyin}, journal={arXiv preprint arXiv:2102.04306}, year={2021} }

@article{vmunetruan2024vm, title={Vm-unet: Vision mamba unet for medical image segmentation}, author={Ruan, Jiacheng and Xiang, Suncheng}, journal={arXiv preprint arXiv:2402.02491}, year={2024} }

@article{mambaunetwang2024mamba, title={Mamba-unet: Unet-like pure visual mamba for medical image segmentation}, author={Wang, Ziyang and Zheng, Jian-Qing and Zhang, Yichi and Cui, Ge and Li, Lei}, journal={arXiv preprint arXiv:2402.05079}, year={2024} }

@article{luo2023cfdbench, title={Cfdbench: A comprehensive benchmark for machine learning methods in fluid dynamics}, author={Luo, Yining and Chen, Yingfa and Zhang, Zhen}, year={2023}, publisher={Preprints} }

@book{fvmeymard2000finite, title={Finite volume methods}, author={Eymard, Robert and Gallou{"e}t, Thierry and Herbin, Rapha{`e}le}, journal={Handbook of numerical analysis}, volume={7}, pages={713--1018}, year={2000}, publisher={Elsevier} }

@book{fdmsmith1985numerical, title={Numerical solution of partial differential equations: finite difference methods}, author={Smith, Gordon D}, year={1985}, publisher={Oxford university press} }

@book{femzienkiewicz2005finite, title={The finite element method set}, author={Zienkiewicz, Olek C and Taylor, Robert L}, year={2005}, publisher={Elsevier} }

@article{pinnsraissi2019physics, title={Physics-informed neural networks: A deep learning framework for solving forward and inverse problems involving nonlinear partial differential equations}, author={Raissi, Maziar and Perdikaris, Paris and Karniadakis, George E}, journal={Journal of Computational physics}, volume={378}, pages={686--707}, year={2019}, publisher={Elsevier} }

@article{mupinnzhang2023multi, title={Multi-Viscosity Physics-Informed Neural Networks for Generating Ultra High Resolution Flow Field Data}, author={Zhang, Sen and Guo, Xiao-Wei and Li, Chao and Zhao, Ran and Yang, Canqun and Wang, Wei and Zhong, Yanxu}, journal={International Journal of Computational Fluid Dynamics}, volume={37}, number={4}, pages={279--297}, year={2023}, publisher={Taylor amp; Francis} }

@inproceedings{nudt1yan2023st, title={ST-PINN: A Self-Training Physics-Informed Neural Network for Partial Differential Equations}, author={Yan, Junjun and Chen, Xinhai and Wang, Zhichao and Zhoui, Enqiang and Liu, Jie}, booktitle={2023 International Joint Conference on Neural Networks (IJCNN)}, pages={1--8}, year={2023}, organization={IEEE} }

@article{ma2024u, title={U-mamba: Enhancing long-range dependency for biomedical image segmentation}, author={Ma, Jun and Li, Feifei and Wang, Bo}, journal={arXiv preprint arXiv:2401.04722}, year={2024} }

@article{me, title={A pseudo-time stepping and parameterized physics-informed neural network framework for Navier--Stokes equations}, author={Zhang, Zhuo and Xiong, Xiong and Zhang, Sen and Wang, Wei and Yang, Xi and Zhang, Shilin and Yang, Canqun}, journal={Physics of Fluids}, volume={37}, number={3}, year={2025}, publisher={AIP Publishing} }

@inproceedings{icassp1DBLP:conf/icassp/ParkGYS25,
  author       = {Yongsung Park and
                  Peter Gerstoft and
                  Seunghyun Yoon and
                  Woojae Seong},
  title        = {Physics-Informed Neural Networks for Ocean Acoustic Field Prediction
                  with Envelope Smoothing},
  booktitle    = {2025 {IEEE} International Conference on Acoustics, Speech and Signal
                  Processing, {ICASSP} 2025, Hyderabad, India, April 6-11, 2025},
  pages        = {1--5},
  publisher    = {{IEEE}},
  year         = {2025},
  url          = {https://doi.org/10.1109/ICASSP49660.2025.10888842},
  doi          = {10.1109/ICASSP49660.2025.10888842},
  bibsource    = {dblp computer science bibliography, https://dblp.org}
}

@inproceedings{icassp2DBLP:conf/icassp/Zhang0025,
  author       = {Gaowei Zhang and
                  Wei Wang and
                  Yi Wang},
  title        = {TIDE-Net: {A} Physics-Based Graph Model for Predicting Tropical Cyclone
                  Impacts on Estuarine Systems},
  booktitle    = {2025 {IEEE} International Conference on Acoustics, Speech and Signal
                  Processing, {ICASSP} 2025, Hyderabad, India, April 6-11, 2025},
  pages        = {1--5},
  publisher    = {{IEEE}},
  year         = {2025},
  url          = {https://doi.org/10.1109/ICASSP49660.2025.10887952},
  doi          = {10.1109/ICASSP49660.2025.10887952},
  bibsource    = {dblp computer science bibliography, https://dblp.org}
}

@article{nudtxinhaichen2024neural,
  title={A neural network approach for unstructured mesh quality evaluation},
  author={Chen, Xinhai and Wang, Zhichao and Liu, Yang and Pang, Yufei and Chen, Bo and Chen, Jianqiang and Gong, Chunye and Liu, Jie},
  journal={Engineering Computations},
  year={2024},
  publisher={Emerald Publishing Limited}
}

@inproceedings{aaai5yang2023dmis,
  title={Dmis: Dynamic mesh-based importance sampling for training physics-informed neural networks},
  author={Yang, Zijiang and Qiu, Zhongwei and Fu, Dongmei},
  booktitle={Proceedings of the AAAI Conference on Artificial Intelligence},
  volume={37},
  number={4},
  pages={5375--5383},
  year={2023}
}

@inproceedings{nips1DBLP:conf/nips/HwangL24,
  author       = {Youngsik Hwang and
                  Dong{-}Young Lim},
  editor       = {Amir Globersons and
                  Lester Mackey and
                  Danielle Belgrave and
                  Angela Fan and
                  Ulrich Paquet and
                  Jakub M. Tomczak and
                  Cheng Zhang},
  title        = {Dual Cone Gradient Descent for Training Physics-Informed Neural Networks},
  booktitle    = {NeurIPS 2024, Vancouver,
                  BC, Canada, December 10 - 15, 2024},
  year         = {2024},
  url          = {http://papers.nips.cc/paper\_files/paper/2024/hash/b2b781badeeb49896c4b324c466ec442-Abstract-Conference.html},
  bibsource    = {dblp computer science bibliography, https://dblp.org}
}

@inproceedings{nips3DBLP:conf/nips/DangelMZ24,
  author       = {Felix Dangel and
                  Johannes M{\"{u}}ller and
                  Marius Zeinhofer},
  editor       = {Amir Globersons and
                  Lester Mackey and
                  Danielle Belgrave and
                  Angela Fan and
                  Ulrich Paquet and
                  Jakub M. Tomczak and
                  Cheng Zhang},
  title        = {Kronecker-Factored Approximate Curvature for Physics-Informed Neural
                  Networks},
  booktitle    = {NeurIPS 2024, Vancouver,
                  BC, Canada, December 10 - 15, 2024},
  year         = {2024},
  url          = {http://papers.nips.cc/paper\_files/paper/2024/hash/3d27d607586984908900eaa8ce19c96c-Abstract-Conference.html},
  bibsource    = {dblp computer science bibliography, https://dblp.org}
}

@inproceedings{icml1DBLP:conf/icml/SalehGBO024,
  author       = {Ehsan Saleh and
                  Saba Ghaffari and
                  Timothy Bretl and
                  Luke N. Olson and
                  Matthew West},
  title        = {Learning from Integral Losses in Physics Informed Neural Networks},
  booktitle    = {{ICML} 2024,
                  Vienna, Austria, July 21-27, 2024},
  year         = {2024},
  url          = {https://openreview.net/forum?id=itDhUBY2xf},
  bibsource    = {dblp computer science bibliography, https://dblp.org}
}

@inproceedings{icml2DBLP:conf/icml/ChoJLL00P24,
  author       = {Woojin Cho and
                  Minju Jo and
                  Haksoo Lim and
                  Kookjin Lee and
                  Dongeun Lee and
                  Sanghyun Hong and
                  Noseong Park},
  title        = {Parameterized Physics-informed Neural Networks for Parameterized PDEs},
  booktitle    = { {ICML} 2024,
                  Vienna, Austria, July 21-27, 2024},
  year         = {2024},
  url          = {https://openreview.net/forum?id=n3yYrtt9U7},
  bibsource    = {dblp computer science bibliography, https://dblp.org}
}

@article{me2zhang2025legend,
  title={Legend-KINN: A Legendre Polynomial-Based Kolmogorov-Arnold-Informed Neural Network for Efficient PDE Solving},
  author={Zhang, Zhuo and Xiong, Xiong and Zhang, Sen and Wang, Wei and Zhong, Yanxu and Yang, Canqun and Yang, Xi},
  journal={Expert Systems with Applications},
  pages={129839},
  year={2025},
  publisher={Elsevier}
}
\end{document}